\documentclass{article}

\usepackage[dblblindworkshop, final]{neurips_2026}
\workshoptitle{Trustworthy AI for Good (AI4GOOD)}

\usepackage[utf8]{inputenc} 
\usepackage[T1]{fontenc}    
\usepackage{hyperref}       
\usepackage{url}            
\usepackage{booktabs}       
\usepackage{amsfonts}       
\usepackage{nicefrac}       
\usepackage{microtype}      
\usepackage{xcolor}         
\usepackage{comment}

\title{The Normalization of Deviance in AI Development}

\author{%
  Emilio Barkett$^{1}$\textsuperscript{*}
  \qquad
  Alexander Kimpton$^{2,1}$
  \qquad
  Daniel Graham$^{3,1}$
  \qquad
  Yusuf Kundgol$^{4,1}$
  \\[1em]
  $^1$Societal Impacts Research on AI\quad $^2$PIBBSS\quad
  $^3$State Street\quad
  $^4$Alignment Research Engineer Accelerator (ARENA)
}

\newcommand{\invisfootnote}[1]{%
  \begingroup
  \renewcommand\thefootnote{}%
  \footnotetext{#1}%
  \endgroup
}

\begin{document}

\maketitle

\invisfootnote{\textsuperscript{*}Correspondence to\ \ \url{emilio@sira-project.org}.}

\begin{abstract}

Work on the risks of artificial intelligence has focused predominantly on capability risk: the danger that systems become too powerful, too autonomous, or too misaligned with human values. Far less attention has been paid to the organizational level---to whether the institutions building these systems are themselves predisposed to drift toward failure. This paper argues that they are. Regardless of how capable AI systems become, the organizations building them face the same structural dynamics that preceded past major technological disasters. Drawing on case studies of the Space Shuttle \textit{Challenger}, the Three Mile Island accident, and the Boeing 737 MAX crashes, this paper identifies the common structural mechanisms preceding each failure and maps them onto contemporary AI development. The findings suggest that existing safety infrastructure may provide less protection than it appears, as organizations can complete safety processes in full compliance and still produce catastrophic outcomes. The pre-disaster period of AI development is still underway; the purpose of this paper is to make these dynamics legible while they can still be interrupted.

\end{abstract}

\section{Introduction}
\label{sec:introduction}

Concerns about the risks of intelligent machines are a century and a half old, from Victorian-era essayists to contemporary AI safety researchers, and have taken on sustained urgency since the release of GPT-3.5 in November 2022. Those warnings have focused almost exclusively on capability risk, the danger of systems becoming too powerful, too autonomous, or too misaligned to control. This paper argues that regardless of how capable AI systems become, the organizations building them may be structurally predisposed to drift toward failure, not through negligence or malice, but through organizational dynamics that have historically preceded major technological disasters.

The implication is that existing AI safety infrastructure---internal safety teams, red-teaming exercises, model evaluations, and responsible scaling policies---is not the protection it appears, as organizations can complete every safety process and \emph{still} drift toward failure because the dynamics identified here operate through those same processes. The argument proceeds as follows. Section~\ref{sec:background} develops the theoretical framework linking organizational structure to failure. Section~\ref{sec:case-studies} turns to history, examining three technological disasters to isolate the structural pattern common to each. Section~\ref{sec:normalization-ai} maps that pattern onto AI development, showing where the same dynamics are already operating. Section~\ref{sec:recommendations} closes with recommendations aimed at the gap this pattern exposes. Section~\ref{sec:limitations} addresses the limits of the historical analogy, Section~\ref{sec:future-work} outlines directions for empirical extension, and Section~\ref{sec:conclusion} concludes.

\section{Background}
\label{sec:background}

Merton observed as early as 1936 that any system of purposive social action inevitably generates unanticipated consequences that run counter to its objectives \cite{merton-1936-unanticipated}. Building on this, Turner identified ``failures of foresight'' as the common precursor to large-scale disasters, arguing that early warning signs were routinely overlooked because of cultural beliefs about hazards and their avoidance \cite{turner-1978-manmade}. Hughes extended this insight to large-scale sociotechnical systems, showing how assemblages of actors, technologies, and organizations produce complexity that is beyond the intentions of any single designer \cite{hughes-1979-electrification, hughes-1993-networks}. Perrow drew the sharpest conclusion from this tradition, arguing that in tightly coupled, complex systems, catastrophic failure is not an anomaly but a structural inevitability built into the system's architecture \cite{perrow-1984-normal}. Sagan broadened the frame, showing how political pressure and bureaucratic interest within larger institutional contexts shape the behavior of organizational units in ways that compound these dynamics \cite{sagan-1993-limits}. Vaughan synthesized and extended this tradition by centering organizational culture and social structure, explaining not just how failures happen but why the people closest to them so often fail to see them coming \cite{Vaughan-1996-Challenger}. Dekker added a temporal dimension, describing how organizations under sustained competitive pressure drift incrementally toward the boundary of safe operation, with each small adaptation making the next one easier to rationalize \cite{dekker-2011-drift}. Together, these contributions locate the origin of catastrophic failure in the ordinary dynamics of organizations under commercial and political pressure.

Out of this tradition emerges the framework central to the present analysis. This process, termed the \textit{normalization of deviance}, describes the incremental organizational drift by which a practice that violates safety norms becomes, through repeated non-disaster, redefined as normal---a concept developed through a landmark study of the 1986 Space Shuttle \textit{Challenger} disaster \cite{Vaughan-1996-Challenger}. A critical feature of this framework is that normalization of deviance does not require villains. The engineers and managers who authorized the launch were competent, experienced, and they sincerely believed the launch was safe. Their belief was wrong, but it emerged from an organizational context that had systematically supplied them with information and frameworks that supported it \cite{vaughan-1999-dark}. This stands in contrast to the ``amoral calculator'' model of organizational failure \cite{carroll-1987-search, kagan1984criminology}, in which decision-makers consciously weigh the costs and benefits of violating safety norms and choose to proceed when benefits outweigh costs \cite{vaughan-1999-dark}. Normalization of deviance does not originate in individual calculation of this kind. Rather, it originates in organizational structure and the culture that structure sustains.

The organizational failure literature identifies four mechanisms that reliably appear across instances of normalized deviance \cite{Vaughan-1996-Challenger, vaughan-1999-dark, dekker-2011-drift}. \textit{Production pressure} creates incentives to minimize the friction that safety processes impose, progressively shifting the burden of proof  onto those arguing for caution. \textit{False assurance from prior success} treats the absence of disaster as evidence of safety, a logically invalid inference that is nonetheless organizationally compelling. \textit{Structural secrecy} describes how organizational structure itself impedes the flow of safety-relevant information through the natural consequences of division of labor and hierarchical reporting \cite{vaughan-1999-dark}. \textit{Erosion of independent oversight} occurs as the relationship between regulators and regulated industries becomes, over time, less adversarial and more collaborative, vitiating the independence that gives oversight its value.

These mechanisms apply with particular force to emerging technologies, where safety standards are developed alongside deployment, with no stock of operational experience to derive them from \cite{vaughan2004-theorizing}. A systematic review of thirty-three studies across oil and gas, nuclear, aviation, healthcare, and rail industries found consistent evidence of these dynamics in every sector examined, suggesting they belong to the structure of organizations managing risk under competitive pressure, whatever the domain \cite{sedlar-2023-systematic}. In mature technological domains, norms are built up over decades, often at the cost of prior accidents. In an emerging technology, this process is foreshortened, or absent entirely. Technologies are easiest to control before their effects are understood, but their effects are not understood until they are widely deployed \cite{collingridge-1980-social, genus-2018-collingridge}. With no mature standard to measure against, industry practice becomes the baseline \cite{david-1985-clio, arthur-1989-competing}. This is the condition in which AI development currently finds itself.

\section{Case Studies}
\label{sec:case-studies}

The cases were selected for their records. Each is among the most exhaustively investigated
organizational failures of the past half-century, and that documentation is what the study of
normalization requires and what intact organizations withhold. The cases span independent
industries and decades, and each foregrounds a different mechanism. They establish how the
dynamics operate, not how often they end in failure.

\subsection{The Challenger Launch Decision (1986)}

On the night of January 27, 1986, engineers from Morton Thiokol conducted an emergency teleconference with NASA administrators. They argued against the following morning's launch, citing the below-freezing forecast and a concern that the O-ring seals in the Solid Rocket Boosters could fail. Their concerns were overruled. Hours later, \textit{Challenger} lifted off, and 73 seconds into the flight, it broke apart, killing all seven astronauts \cite{higginbotham-2024-challenger, rogers-1986-report, mcdonald-2012-truth}.

The concept of the normalization of deviance emerged directly from analysis of this disaster \cite{Vaughan-1996-Challenger}. In the years leading up to \textit{Challenger}, engineers had openly observed and documented O-ring erosion, but each flight that survived without catastrophe was read, retrospectively, as confirming the safety case. Over time, the organizational interpretation of this evidence shifted. Production pressure, driven by a launch schedule that had already slipped multiple times, had progressively reversed the burden of proof: where safety once had to be established before flight, those urging caution now had to prove the launch unsafe in order to halt it. False assurance from prior success closed the argument. Each uneventful flight had made the next approval easier.

\subsection{The Accident at Three Mile Island (1979)}

The partial meltdown at Three Mile Island Unit 2 on March 28, 1979, began with a stuck-open pilot-operated relief valve in a reactor system that had accumulated years of minor anomalies, each handled through improvised workarounds that had gradually calcified into standard practice \cite{walker-2004-three, perrow-1981-normal}. The President's Commission concluded that the disaster resulted from deficiencies in how the Nuclear Regulatory Commission and the nuclear industry had approached reactor safety \cite{kemeny-1979-report}. The organizational environment had normalized the management of complexity through improvisation, and that normalization was what made the accident possible \cite{perrow-1981-normal}. Three Mile Island is most often read through normal accident theory, as a failure of interactive complexity and tight coupling \cite{perrow-1984-normal}. The reading here is complementary as the informal workaround culture that preceded the accident is what normalization of deviance looks like in an operating environment.

In an organizational context, Three Mile Island foregrounds structural secrecy and opacity. The reactor systems had grown in complexity as modifications were layered on modifications, and operators in the control room could not, in real time, form an accurate mental model of what was happening inside the reactor. When the relief valve stuck open, a cascade of confusing and partially contradictory indicator readings meant consequential decisions were made on an incorrect understanding of the system's state. Operators had been trained to handle the anomalies they expected, and the combination of anomalies that actually occurred fell entirely outside that training \cite{perrow-1981-normal, perrow-1984-normal, walker-2004-three}.

Operators had developed habits of handling routine anomalies informally, without systematic reporting or analysis, because treating every anomaly as a potential crisis was organizationally unsustainable. The informal workarounds worked until they did not. By the time the accident occurred, the gap between formal operating procedures and actual operational practice was wide enough that operators lacked the procedural and cognitive resources to recognize and respond correctly to what was happening. Safety-relevant information existed within the system. It simply never reached those with the authority and knowledge to act on it.

\subsection{The Boeing 737 MAX Crashes (2018--2019)}

The crashes of Lion Air Flight 610 in October 2018 and Ethiopian Airlines Flight 302 in March 2019 were caused by a software system called MCAS (Maneuvering Characteristics Augmentation System) that repeatedly pushed the nose of the aircraft down in response to faulty sensor data \cite{robison-2022-flying}. MCAS was designed to compensate for aerodynamic handling differences created by installing larger engines on an airframe not designed to accommodate them. It was a software patch for a physical design flaw, and the pilots of both flights were unable to override the system before impact.

The origins of this design trajectory lie in competitive pressure \cite{johnston-2019-boeing, macarthur-2020-cost}. When Airbus announced the A320neo, Boeing faced a choice: design a new aircraft or modify the existing 737 platform. Modification was faster and cheaper, and would allow the aircraft to be certified as a variant of the existing 737 rather than as an entirely new airplane, avoiding costly simulator training requirements for airline customers \cite{house-2020-737max, hopkins-2025-boeing}. Boeing chose modification and committed to a path in which a series of engineering compromises became increasingly difficult to walk back. MCAS was initially a minor system designed to engage only in rare high-angle-of-attack situations (nose pitched steeply upward). Over the course of development, its authority was expanded and it was made to rely on input from a single angle-of-attack sensor, making it vulnerable to a single-point failure. Each change was accompanied by an assessment that the modification was acceptable \cite{herkert-2020-boeing}.

The FAA's role illustrates erosion of independent oversight in its most developed form. Boeing employees acting as FAA Authorized Representatives conducted much of the safety analysis and certification work that the FAA nominally oversaw \cite{house-2020-737max}. The institutional trust that had developed between Boeing and the FAA over decades of generally successful collaboration had become a substitute for independent verification \cite{robison-2022-flying}. The Congressional investigation found that FAA managers had overruled their own engineers when those engineers raised concerns about the certification process \cite{house-2020-737max}. The organizations nominally responsible for preventing normalization had become participants in it.

\section{Normalization of Deviance in AI Development}
\label{sec:normalization-ai}

The three cases examined above share a common structural pattern: an extended period of drift during which warning signals mounted and were progressively redefined as acceptable, enabled by production pressure, false assurance from prior success, structural secrecy, and erosion of independent oversight. Each mechanism is observable in contemporary AI development, and several features of the AI context make the conditions more acute than in prior technological domains and also compounded. Mechanisms that appeared singly in the historical cases appear together here. NASA drifted inside mature procedures, and Boeing certified against five decades of experience with the 737; the frontier laboratories are adolescent institutions, none much more than a decade old, operating all four mechanisms at once on systems less legible than any their predecessors managed.

\textbf{Production pressure} in AI development---commercial competition among a handful of frontier laboratories, overlaid by strategic rivalry between the United States and China---operates at an intensity that has few historical precedents outside of wartime \cite{armstrong-2016-racing, rhodes-1986-atomic, sagan-1993-limits}. The economic and reputational stakes attached to being first create sustained pressure to treat safety work as friction, not value \cite{felstead-2025-frontier, armstrong-2016-racing, cave-2018-race}. Safety evaluations that might delay a release, red-teaming exercises whose findings might block a product, and interpretability research that reveals concerning model behaviors each represent a potential impediment to competitive position. The burden of proof has shifted in a manner structurally identical to what occurred at NASA: those arguing for caution must overcome those arguing for speed, rather than the reverse \cite{armstrong-2016-racing, felstead-2025-frontier, bostrom-2014-superintelligence, cave-2018-race}. At the state level, AI capability has been explicitly framed as a matter of national security and economic survival \cite{state-council-2017-plan, biden-2024-nsm, horowitz-2018-balance}.

\textbf{False assurance from prior success} manifests in AI development through benchmark performance and deployment history \cite{raji-2020-closing, bengio-2025-safety, stanovsky-2025-benchmarks}. AI systems are assessed using standardized evaluation suites, and performance on those benchmarks is treated as safety evidence \cite{liang-2023-helm, hendrycks-2021-unsolved}. But the history of AI benchmarks is a history of systems that perform well on evaluations while exhibiting concerning behaviors the benchmarks were not designed to detect \cite{raji-2020-closing, ribeiro-2020-beyond}. When a model passes its evaluations and is deployed without incident, this is treated as validation of both the model and the evaluation framework. Each successive generation of AI systems that is deployed without catastrophe expands the implicit safety baseline for the next, making it progressively harder to argue for more stringent standards as capabilities increase \cite{bommasani-2021-opportunities}. This is particularly dangerous as we begin to uncover increasing evidence of frontier models scheming on a much larger scale than we had previously realized \cite{openai2026hf}. This concern is no longer hypothetical. OpenAI's report on the July 2026 Hugging Face incident documents evaluation-stage agents escaping containment through chained exploits and compromising third-party production infrastructure \cite{openai2026hf}.

\textbf{Structural secrecy} follows from the ordinary organization of frontier AI laboratories \cite{vaughan-1999-dark, anderljung-2023-frontier, cihon-2025-auditing}. Safety teams, product teams, and leadership operate in different organizational contexts with different incentive structures and different information environments \cite{delaney-2024-mapping, cihon-2025-auditing}. Concerns raised by safety researchers may not reach or influence deployment decisions made under competitive pressure. The pattern is documented across multiple frontier organizations in the public departures of safety-focused researchers citing the prioritization of products over safety \cite{goldman-2024-leike, techbrew-2026-departures, ethicalai-2024-departures}. Red-team findings may be documented without carrying the authority to change outcomes \cite{ganguli-2022-red, perez-2022-red}. The consequence of division of labor and hierarchical reporting is that safety-relevant information fails to reach those with authority to act on it.

\textbf{Erosion of independent oversight} is perhaps the most advanced of the four mechanisms in AI development \cite{rost-2026-disclosure, ho-2023-international, anderljung-2023-frontier}. The governance infrastructure for AI is embryonic by historical standards, and the organizations with the strongest interest in less stringent standards have the greatest influence over how those standards are set \cite{anderljung-2023-frontier, ho-2023-international, wei-2024-finetune}. The expertise needed to evaluate frontier systems is generated by the work of building them, so it accumulates where development happens and nowhere else. Every year of frontier progress widens the gap between what laboratories can assess and what any outside body can \cite{cihon-2025-auditing}. Responsible scaling policies and model evaluations are defined, conducted, and enforced by the same organizations whose deployment decisions they are meant to constrain \cite{coggins-2025-prepare, rost-2026-disclosure, openai-2023-preparedness, openai-2025-preparedness}. This is the self-certification dynamic that the Congressional investigation identified as a central institutional cause of the 737 MAX crashes \cite{robison-2022-flying, house-2020-737max}, and it is the current default governance model for frontier AI development \cite{hadfield-2023-regulatory, anderljung-2023-frontier, rost-2026-disclosure}. When the entity defining safety standards has an institutional stake in the content of those standards, normalization of deviance can proceed regardless of individual or institutional good faith \cite{anderljung-2023-frontier}.

\section{Recommendations}
\label{sec:recommendations}

The preceding analysis suggests that normalization of deviance in AI development is an organizational, cultural, governance, and technical problem. Because normalization operates through rule-following rather than rule-breaking, interventions that merely add new rules, compliance requirements, or accountability mechanisms are unlikely to be sufficient on their own. The recommendations below are organized around the four mechanisms identified above, with each designed to address a structural condition.

\textbf{Organizational.} The common thread in all three case studies is that safety functions lost practical independence from production incentives. At NASA, safety engineers reported to the same management chain under pressure to launch \cite{Vaughan-1996-Challenger}. At the FAA, the self-certification model embedded the certifying authority within the organization whose products it was certifying \cite{robison-2022-flying}. Safety functions require structural independence from the processes that create production pressure. In AI development, this implies safety organizations that report through structures insulated from deployment timelines, with authority to delay or halt deployment that is not subject to override by the same management chain under pressure to ship \cite{cihon-2025-auditing, Vaughan-1996-Challenger}. It also implies that safety findings are reported externally as well as internally, so that structural secrecy cannot operate unchecked. Pre-mortem analysis conducted before deployment directly addresses the false assurance mechanism. Requiring organizations to prospectively construct the plausible chain of decisions that could produce failure forces implicit risk tolerances to become explicit \cite{klein-2007-premortem}.

\textbf{Epistemic and cultural.} Organizations cannot defend against normalization of deviance if they do not recognize it as a distinct phenomenon. The amoral calculator model attributes failures to bad actors making bad calculations, providing a more comfortable explanation than the normalization of deviance account \cite{Vaughan-1996-Challenger, kagan1984criminology}. Naming the dynamic explicitly and incorporating it into the safety culture and training of AI development organizations is a prerequisite for the other interventions to function \cite{vaughan-1999-dark, sedlar-2023-systematic}. Concretely, this means building epistemic cultures in which prior successful deployments are treated as precedent rather than proof. Each new generation of more capable systems requires its own affirmative safety evidence, independent of the deployment history that preceded it \cite{Vaughan-1996-Challenger, dekker-2011-drift}. It means treating anomalous model behaviors as meaningful signals, with the institutional expectation that they are systematically investigated, and not absorbed into an existing risk category \cite{vaughan-1999-dark, sedlar-2023-systematic}. The institutional protection of dissenting voices deserves special emphasis. In all three cases examined, individuals within the organization raised concerns through appropriate channels and were overruled through appropriate processes. The organizational structure lacked the capacity to give dissent due weight. Mechanisms that give safety concerns procedural standing, requiring that safety objections be formally documented, responded to in writing, and escalated independently of the management chain under production pressure, address this structural failure directly \cite{ganguli-2022-red, delaney-2024-mapping}.

\textbf{Regulatory and governance.} The 737 MAX case is the clearest illustration of why industry self-regulation, however well-intentioned, is structurally insufficient to prevent normalization. Boeing's engineers knew the risks, their safety processes identified concerns, and internal culture systematically overrode those concerns under commercial pressure. The FAA's self-certification regime provided the institutional framework within which this override could proceed with apparent legitimacy. Strong internal safety culture is necessary but requires external, independent oversight to function as intended. For AI development, this implies third-party evaluation infrastructure that does not depend on developer cooperation for access, funded through sources other than the organizations it evaluates, and empowered to conduct and publish its own assessments of safety-relevant model properties \cite{anderljung-2023-frontier, ho-2023-international, wei-2024-finetune, cihon-2025-auditing}. The National Transportation Safety Board (NTSB) model is instructive. Its investigations are independent of the FAA and of industry, and its findings are published in full. An AI analog does not need to replicate every feature of the NTSB to break the secrecy that lets normalization proceed inside the closed loop of industry self-assessment. Incident reporting systems modeled on the Aviation Safety Reporting System (ASRS), anonymous, no-fault, and focused on systemic learning, directly address the same dynamic, creating channels through which safety-relevant information flows toward analysis instead of remaining captured where it was produced \cite{hadfield-2023-regulatory, ho-2023-international}.

\textbf{Technical.} Both the Three Mile Island case and Section \ref{sec:normalization-ai} identify opacity as a primary driver of normalization: when operators cannot see the system, improvisation fills the gap, and improvisation normalizes. The TMI operators could not form an accurate mental model of the reactor's state; AI safety evaluators cannot verify that the properties producing safe behavior in test conditions will produce safe behavior in deployment. Interpretability research should therefore be treated as a safety prerequisite and funded accordingly \cite{hendrycks-2021-unsolved, bengio-2025-safety}. Staged deployment with enforceable rollback criteria addresses the deployment normalization dynamic. Staged deployment that treats advancing to each stage as the default and halting as requiring affirmative evidence of failure replicates the burden-of-proof reversal that was the defining symptom of normalization at NASA. Conditionally staged deployment, in which advancing requires affirmative evidence of safety and the authority to halt is structurally independent from the authority to ship, confronts the same mechanism that the \textit{Challenger} pre-launch review failed to address \cite{shevlane-2023-model, Vaughan-1996-Challenger}.

\section{Limitations}
\label{sec:limitations}

The argument advanced here depends on a structural analogy between AI development organizations and the organizations that produced the \textit{Challenger}, Three Mile Island, and Boeing 737 MAX disasters. This analogy is the paper's central contribution, but it is not without weaknesses.

The most fundamental limitation is epistemic, and it is invidious---the framework's full evidentiary record only ever exists after the failure it predicts. The dynamic is invisible to those undergoing it, and disaster is what makes the accumulation of small decisions visible. This paper therefore cannot offer a smoking gun. The structural conditions identified here are consistent with normalization of deviance occurring in AI development today, but consistency is not proof. The documented departures of safety researchers, the revision histories of responsible scaling policies, and the gap between benchmark performance and deployment safety are each consistent with normalization---they are also consistent with organizations learning and adapting under difficult conditions. Distinguishing between these interpretations requires evidence that is not yet available. One partial discriminator does exist: learning and normalization predict different directional patterns across a growing corpus of policy revisions, with learning tracking new safety information and normalization tracking competitive events.

A second limitation concerns the boundaries of the analogy. The normalization of deviance framework was developed for physical engineering systems where failure modes are relatively legible (e.g., an O-ring can be inspected, erosion can be measured, and a causal chain can be reconstructed after the fact). The failure modes of AI systems are more diffuse, the relevant properties less measurable, and the distinction between a concerning anomaly and normal system behavior often simply unclear. If anything, the disanalogy heightens the concern, since what cannot be seen cannot be corrected. But the mapping is imperfect, and the framework has to be applied with that in view.

A third limitation is that the AI safety community is substantially more self-aware about these risks than NASA, the NRC, or Boeing were at comparable stages. The field of AI safety exists as a recognized discipline; major development organizations employ substantial safety teams; the intellectual frameworks for thinking about misalignment, evaluation failure, and organizational risk have been developed and published widely. Several of the laboratories responsible for these risks are also their most prominent public chroniclers and advocates---safety commentary is, for some, part of the commercial identity. This paper does not claim that AI development organizations are indifferent to safety. The argument is that awareness of a risk is insufficient to prevent normalization when the organizational structures that produce it remain intact, and this distinction is easier to assert than to verify empirically.

Finally, the paper conflates corporate-level and state-level AI development to some degree, treating them as instances of the same phenomenon operating at different scales. The organizational dynamics and incentive structures relevant to a commercial AI laboratory differ substantially from those relevant to a state-level AI program. A fuller treatment would separate the two and work out the governance implications of each.

\section{Future Work}
\label{sec:future-work}

The argument advanced here is necessarily preliminary. To the authors' knowledge, this is the first sustained application of the normalization of deviance literature to frontier AI development, and its primary contribution is conceptual rather than empirical. The historical cases demonstrate the organizational dynamics that produce normalization, but do not by themselves establish that the same process is occurring within AI organizations. Future work should therefore be empirical. Three approaches seem most tractable. 

First, longitudinal analysis of frontier laboratories' governance artifacts, including responsible scaling policies, preparedness frameworks, and model cards, treating successive revisions as a corpus whose direction of change can be coded and correlated with competitive events and safety incidences. Under normalization, procedural language would be expected to become more permissive over time and competitive-adjustment provisions more entrenched; under learning, capability thresholds would be qualified or strengthened. These predictions can be tested against the public revision histories already available.

Second, systematic analysis of personnel testimony. The departures of safety and governance researchers from frontier laboratories between 2024 and 2026, notably Daniel Kokotajlo, Jan Leike, and William Saunders, have produced an unusually public record of first-hand accounts, several of which describe precisely the pattern of interest: standards eroding through the accumulation of deployments in which nothing visibly went wrong \cite{kokotajlo-2024, leike-2024, saunders-2024-testimony}. Treated as evidence in aggregate and read longitudinally, these accounts can be situated on the same timeline as governance artifact revisions and competitive events, with convergence across independent accounts corroborating the normalization interpretation. 

Third, extension to the state level. Nation-states competing over AI capability are themselves organizations under production pressure, and the framework applied to them directly \cite{sagan-1993-limits, horowitz-2018-balance}. National strategies, export control rules, and procurement and testing standards are published and revised in ways tractable to the same analysis proposed above, and the comparative question of whether safety commitments erode differently under state principals than under market discipline would begin to separate the corporate and state cases this paper has treated together.

\section{Conclusion}
\label{sec:conclusion}

The normalization of deviance is a story about ordinary people, following ordinary procedures, inside well-intentioned organizations, collectively drifting toward catastrophic failure. Analysis of the \textit{Challenger} disaster revealed that the organizational conditions producing catastrophe are structural---the predictable output of institutions managing risk under competitive pressure without adequate independence for their safety functions \cite{Vaughan-1996-Challenger}. Three Mile Island and the Boeing 737 MAX confirmed that this dynamic generalizes across industries and decades. This paper has argued that the structural conditions enabling normalization of deviance are present in AI development in acute form. Production pressure, false assurance from prior success, structural secrecy, and erosion of independent oversight are each observable in AI development today.

The opacity of AI systems makes warning signals harder to read, and the immaturity of the regulatory environment leaves safety norms to be written by the developers themselves. Competitive pressure at both the corporate and state levels creates sustained incentives to treat safety work as friction. Practices that would register as deviant against a mature standard simply become the standard. The disasters examined in this paper were catastrophic. They were also legible: investigators could determine what had happened, attribute it to organizational causes, and construct institutional reforms designed to prevent recurrence. A sufficiently large AI safety failure may produce consequences that are neither recoverable nor legible enough to serve as a corrective, foreclosing the institutional learning that followed prior disasters.

The AI development community is in the period analogous to the years before the major accidents examined here, when norms are still being written and institutions are still being designed. The organizational, cultural, regulatory, and technical recommendations offered here share a common logic: safety infrastructure built before competitive pressures reach their peak is more robust than infrastructure constructed in response to failure. Whether the current moment is used as an opportunity depends on whether the dynamics identified here are taken seriously by those best positioned to act on them.

\bibliography{refs}
\bibliographystyle{plain} 

\end{document}